\documentclass[sigconf, screen, review=false, natbib=false, anonymous=false]{acmart}
\usepackage{threeparttable}
\usepackage{tcolorbox}
\usepackage{graphicx}
\usepackage{subcaption}
\usepackage{bbding}
\usepackage{multirow}
\usepackage{hyperref}
\usepackage{balance}
\usepackage[table,xcdraw]{xcolor}
\usepackage{makecell} 
\usepackage{fontawesome5} 
\usepackage{algorithmic}
\usepackage{algorithm}
\copyrightyear{2026}
\acmYear{2026}
\setcopyright{acmlicensed}

\acmDOI{}

\RequirePackage[
  datamodel=acmdatamodel,
  style=acmnumeric,
  ]{biblatex}

\bibliography{sample-base.bib}

\begin{document}

\title{An Effective Router for Vision-Language Model Selection}


\author{Can Wang}
\affiliation{%
  \institution{Harbin Institute of Technology}
  \city{Harbin}
  \country{China}
}
\email{23B903072@stu.hit.edu.cn}

\author{Shengwei Wang}
\affiliation{%
  \institution{Harbin Institute of Technology}
  \city{Harbin}
  \country{China}
}
\email{2022211821@stu.hit.edu.cn}

\author{Bolin Zhang}
\authornote{Corresponding Author.}
\affiliation{%
  \institution{Harbin Institute of Technology}
  \city{Harbin}
  \country{China}
}
\email{bolin@hit.edu.cn}

\author{Zhiying Tu}
\authornote{Corresponding Author.}
\affiliation{%
  \institution{Harbin Institute of Technology}
  \institution{Shandong Key Laboratory of Digital Service Computing Technology and Systems}
  \city{Harbin}
  \country{China}
}
\email{tzy\_hit@hit.edu.cn}

\author{Dianhui Chu}
\affiliation{%
  \institution{Harbin Institute of Technology}
  \institution{Shandong Key Laboratory of Digital Service Computing Technology and Systems}
  \city{Harbin}
  \country{China}
}
\email{chudh@hit.edu.cn}

\begin{abstract}

Vision-language models (VLMs) with varying performance and resource requirements are widely deployed, making it difficult for users to select the most appropriate one among numerous VLM candidates. Existing work reveals the performance paradox phenomenon in language models and focuses on routing methods to solve it. However, developing a router for VLM selection is still a critical yet challenging problem, which primarily faces: 1) lack of specialized data, 2) ineffective feature representation, and 3) rigid model space and costly adaptation. 
In this paper, we construct a multimodal dataset for VLM selection, containing the outputs of seven mainstream VLMs on 32,626 unique image-text queries. We then propose ARMS, a router for VLM selection. ARMS enhances input signals with VLM profiles, employs a simple but effective architecture to improve representations of queries and VLM capabilities. To improve ARMS’ adaptation to new VLMs, we propose two extension training strategies: incremental training and independent training. Experimental results on both in-distribution and out-of-distribution test sets demonstrate the effectiveness of ARMS. In particular, using our training strategy, ARMs (only 800M in size) can adapt to a broader VLM space and defeat commercial models like GPT-4o that are hundreds of times larger in scale. Our code, models, and datasets are available in the \href{https://anonymous.4open.science/r/ExpG-F72B}{anonymous repository}.
\end{abstract}

\ccsdesc[500]{Computing methodologies~Computer vision}
\ccsdesc[300]{Computing methodologies~Computer vision representations}
\ccsdesc[300]{Computing methodologies~Machine learning}



\keywords{Model Selection, Vision-Language Models (VLMs), Dataset for VLMs, Router Method}

\maketitle

\section{Introduction}

A variety of models, each with different performance characteristics and resource requirements, are widely used to deliver intelligent responses across diverse domains and applications \cite{app1,app2,app3}. 
Many studies \cite{VHELM,MARS,framework} have revealed a performance paradox in language models: larger and more expensive models do not always outperform their smaller and more efficient counterparts. 
Building on this insight, we conduct an empirical study on the VHELM benchmark \cite{VHELM} and find that the performance paradox also exists among Vision-language models (VLMs). 

\begin{figure}[!t]
  \centering
  \includegraphics[width=\linewidth]{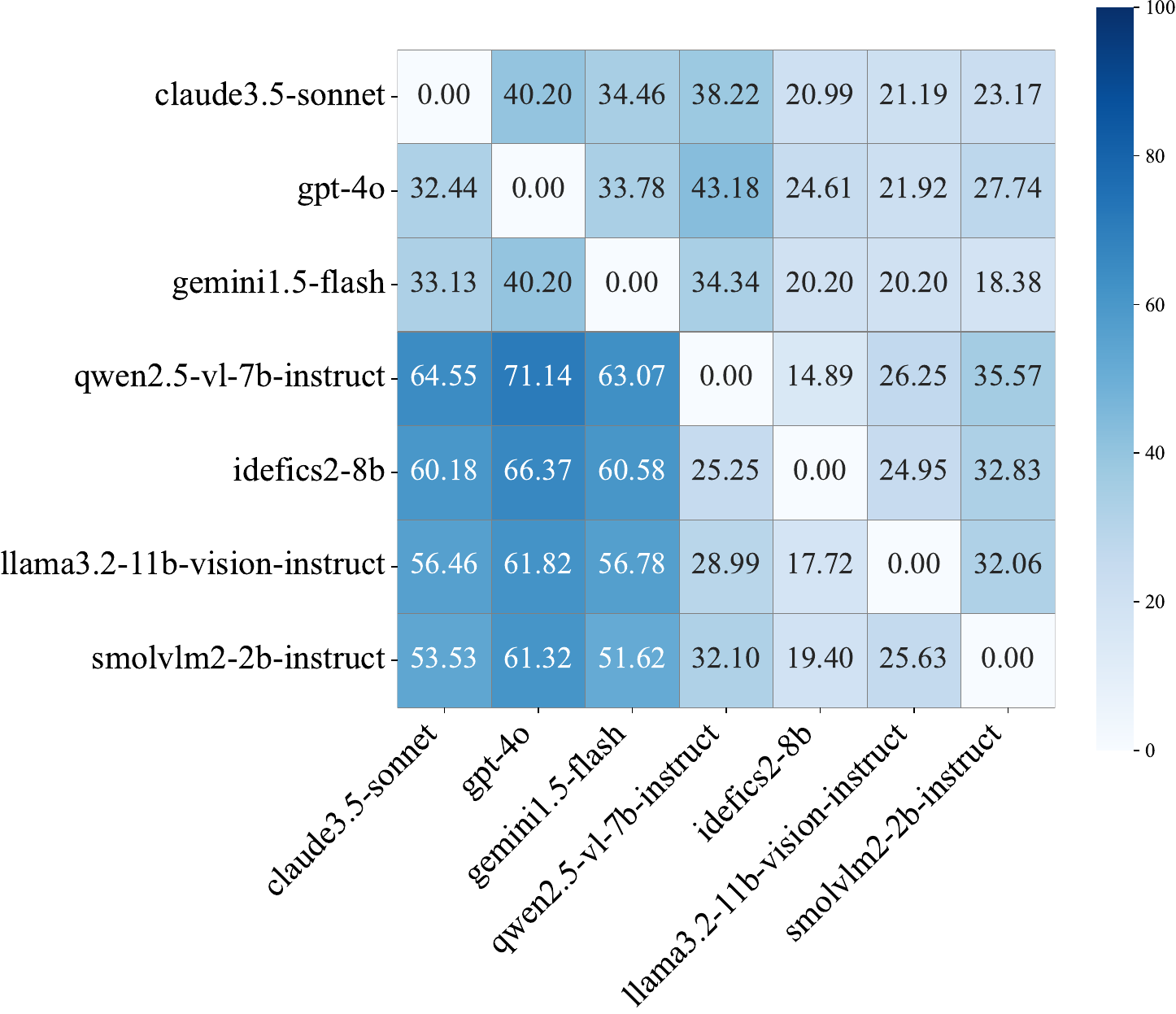}
  \caption{The proportion(\%) of samples answered incorrectly by the row model but correctly by the column model on the VHELM \cite{VHELM} benchmark, showing the performance paradox between VLMs.}
  \Description{Our motivation}
  \label{motivation}
\end{figure}

As shown in Fig.~\ref{motivation}, among the image-text queries that the prominent and paid-access VLM GPT-4o \cite{gpt4o} answered incorrectly, 43.18\% are correctly answered by free and open-source VLM qwen2.5-vl-7b-instruct \cite{qwen2.5}.  Conversely, GPT-4o correctly answered 71.14\% of the queries that Qwen2.5-VL-7B-Instruct failed to handle. The performance of any two VLMs in Fig.~\ref{motivation} is complementary, and no single VLM consistently delivers optimal performance across all tasks and domains. Even lightweight, open-source models can compete with or outperform top-tier models in a significant number of samples.

For real-world applications, the complementary performance of VLMs raises a key question: which model is the most appropriate and best-performing for a given query? 
Existing works \cite{TO-Router, GraphRouter, FORC} develop a router to select models based on different user queries. However, these works fail to adapt to the VLM selection problem, which primarily faces three major obstacles: 
1) Lack of specialized data: existing router research lacks a large-scale multimodal dataset to to train VLM router and fairly assess its performance. 
2) Ineffective feature representation: existing routers typically rely on general-purpose semantic embeddings, which are not specifically designed for VLM selection. As a result, they may fail to fully utilize multimodal inputs and model-related information when making selection decisions, leading to suboptimal performance.
3) Rigid model space and costly adaptation: existing routers are trained over a fixed set of VLMs and learn model-specific decision boundaries, which restricts the selection space to seen models. When new VLMs are introduced, extending the router typically requires retraining over the entire model set, leading to high computational cost and poor scalability in dynamic environments.

To address the above challenges, we propose \textbf{ARMS} (\textbf{a} \textbf{r}outer for VL\textbf{M} \textbf{s}election). To tackle the lack of specialized data, we construct \textbf{M$^2$}, a large-scale multimodal dataset tailored for VLM selection, which provides aligned supervision across diverse image-text queries and multiple VLMs, enabling the router to learn reliable selection signals and supporting comprehensive evaluation. 
To overcome insufficient feature representation for model selection, ARMS adopts a unified architecture that jointly models multimodal inputs and model-related information. In particular, the gate network dynamically routes inputs to different multimodal fusion experts, enabling input-dependent feature transformation. This design allows the model to capture diverse patterns across heterogeneous image-text queries and VLM capabilities, which cannot be effectively handled by a single shared representation. By incorporating structured VLM profiles as capability descriptors and leveraging expert specialization, ARMS learns more informative representations for accurate model selection.
Finally, to address the rigid model space and costly adaptation, we propose two extension training strategies: \emph{incremental training} and \emph{independent training}, these strategies eliminate the need for full retraining over all models and allow ARMS to scale efficiently in dynamic and evolving model ecosystems.

Our contributions are as follows:

\begin{itemize}
\item[1] We release \textbf{M$^2$}, the first large-scale \textbf{m}ultimodal dataset for vision-language \textbf{m}odel selection, covering 56,107 unique image-text queries and 7 VLMs.

\item[2] We propose \textbf{ARMS}, \textbf{a} \textbf{r}outer for VL\textbf{M} \textbf{s}election. ARMS enhances input signals with VLM profiles, employs a simple but effective architecture to improve representations of queries and VLM capabilities, and uses two extension training strategies to enhance its adaptation to unseen VLMs. 

\item[3] Under both in-distribution and out-of-distribution settings, ARMS outperforms all individual VLMs in the initial selection space. With the selection space expanded through two proposed training strategies, ARMS (only 800M) even outperforms bigger commercial model GPT-4o.

\end{itemize}

\section{Related Works}

\textbf{Router Dataset.} Existing works either use HELM \cite{HELM} as a training set or propose their own datasets. HELM conducts a large-scale evaluation of 30 prominent language models (spanning open, limited-access, and closed models) on 42 tasks and has released the output results of each LLM. Table~\ref{tab:dataset} presents a comparison between our dataset (M$^2$) and existing datasets for router training. Although FrugalGPT \cite{FrugalGPT} has publicly released all the data used in its experiments, including 17,982 unique text queries, it does not provide the output results of LLMs, making the dataset unsuitable for router training. While RouterDC \cite{RouterDC}, MARS \cite{MARS}, and HELM \cite{HELM} provide LLM outputs, their queries are purely textual rather than multimodal, rendering them inapplicable to the VLM selection problem. VHELM \cite{VHELM} is a VLM benchmark composed of 21 existing task datasets and randomly samples up to 1,000 instances per task. However, it lacks sufficient samples within individual tasks, making it inadequate for studying the relationship between sample size and router performance in in-distribution settings. Thus, we release the largest dataset for VLM selection known to us so far, M$^2$, containing 56,107 unique multimodal queries.

In summary, existing datasets \cite{RouterDC, MARS, HELM} are suitable only for training routers for LLM selection, while VHELM \cite{VHELM} is applicable for VLM selection. However, VHELM provides too few training samples (fewer than one thousand) per task, making it difficult for the router to accurately assess the true performance of a VLM on a given task with such limited data.

\begin{table}[!t]
  \centering
  \caption{Comparison with existing datasets}
  \resizebox{\linewidth}{!}{
    \begin{tabular}{cccccc}
    \toprule
    Dataset & Multimodal? & Model Outputs? & Target? & Queries\\
    \midrule
    \makecell[c]{FrugalGPT \cite{FrugalGPT}} & \textcolor{red}{\XSolidBrush} & \textcolor{red}{\XSolidBrush} & Selection & 17,982 \\
    \makecell[c]{RouterDC \cite{RouterDC}} & \textcolor{red}{\XSolidBrush} & \textcolor{green!70!black}{\checkmark} & Selection & 8,225 \\
    \makecell[c]{MARS \cite{MARS}} & \textcolor{red}{\XSolidBrush} & \textcolor{green!70!black}{\checkmark} & Selection & 30,000 \\
    \makecell[c]{HELM \cite{HELM}} & \textcolor{red}{\XSolidBrush} & \textcolor{green!70!black}{\checkmark} & Evaluation & 16,000 \\
    VHELM \cite{VHELM} & \textcolor{green!70!black}{\checkmark} & \textcolor{green!70!black}{\checkmark} & Evaluation & 21,000  \\
    M$^2$ (Ours) & \textcolor{green!70!black}{\checkmark} & \textcolor{green!70!black}{\checkmark} & Selection & 56,107  \\
    \bottomrule
    \end{tabular}%
}
  \label{tab:dataset}%
\end{table}%

\textbf{Router Methods.} Existing works are categorized as three types: output-aware methods \cite{FrugalGPT, LLM-Blender}, input-aware Bert-based methods \cite{FORC, TO-Router, Hybrid-LLM} and input-aware other methods \cite{GraphRouter, Routing, ZOOTER}.
Output-aware methods select the optimal output from the responses of multiple LLMs and return it to the user as the final answer to a query. FrugalGPT \cite{FrugalGPT} trains the DistilBERT \cite{DistilBERT} model as a judge to predict the quality of LLMs' answers, and then sequentially invokes a list of LLMs until the judge's score for an answer surpasses a predefined threshold. 
Given a query, LLM-Blender \cite{LLM-Blender} employs PairRanker to distinguish subtle differences between the outputs of candidate LLMs and then use GenFuser to generate an improved output based on the top-ranked candidate outputs. However, these works concurrently invoking multiple LLMs for the same query incurs substantial costs. 

Input-aware BERT-based methods train a BERT-like model \cite{Bert} to determine which LLM performs best, based solely on the user query and the LLM name. FORC \cite{FORC} trains a meta-scorer based on DistilBERT \cite{DistilBERT} to predict the performance score of a LLM given a specific query. TO-Router \cite{TO-Router} trains BERT-Router with soft labels derived from the scaled BERT similarity scores between each query and expert LLM. Hybrid-LLM \cite{Hybrid-LLM} trains DeBERTa-v3-large \cite{DeBERTa} to predict the query difficulty and the desired quality level and then assigns the query to the small or large model. However, these works cannot be extended to unseen LLMs unless retrained, especially as the variety of LLMs continues to grow. Input-aware other methods employ alternative architectures as routers, such as graph-based \cite{GraphRouter}, kNN-based \cite{Routing}, and reward-based \cite{ZOOTER} routers. Their inputs also consist solely of the VLM profile and the text query.


In summary, these existing works are only applicable to scenarios where the user query is purely text-based. When the query includes images, these works fail to capture the information embedded in the images that indicate the user's needs. Moreover, these works do not address the issue of expanding the model selection space, which results in the router's inability to generalize to previously unseen models.

\section{M$^2$ Dataset}

\subsection{Dataset Construction}

In order to construct a comprehensive dataset with sufficient data volume spanning diverse queries, we first selected three large-scale VLM benchmarks, each containing more than 20k unique image-text pairs as queries: ALM-bench \cite{ALM-bench}, MMT-Bench \cite{MMT-Bench}, and MME-RealWorld \cite{MME-RealWorld}. We then run four open-source VLMs, each with a parameter count of no more than 14B, on these queries using 4*A100 servers: qwen2.5-vl-7b-instruct \cite{qwen2.5}, idefics2-8b \cite{idefics2}, llama3.2-11b-vision-instruct \cite{llama3.2}, and smolvlm2-2b-instruct \cite{smolvlm2}. 
After filtering out queries that none of the VLMs could answer correctly, we obtain a total of 32,626 queries, resulting in 130,504 samples (32,626 queries $\times$ 4 VLMs). These samples form the base in-distribution (ID) part in our dataset, which are split into training and test sets for training and evaluating the router.

\begin{figure}[!ht]
  \centering
  \includegraphics[width=0.9\columnwidth]{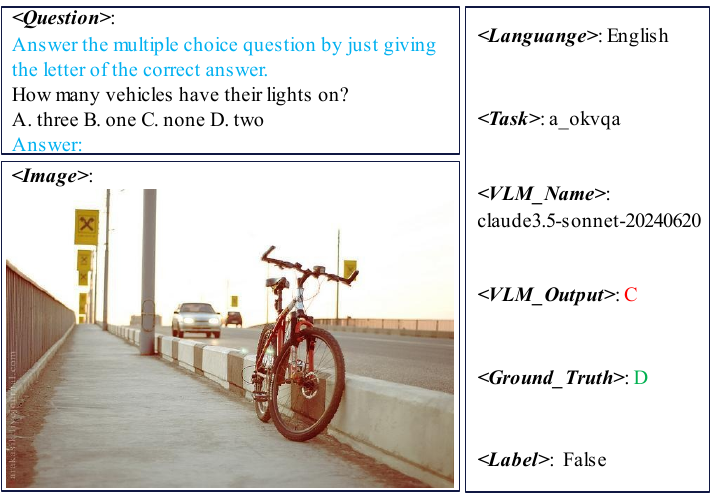}
  \caption{A data sample in M$^2$. $<$Question$>$ and $<$Image$>$ are combined as the input to the VLM claude3.5-sonnet, where the blue text denotes the added prompt for output standardization. The correct answer is in green, while the incorrect answer from the VLM is in red.}
  \Description{Our motivation}
  \label{fig:sample}
\end{figure}

To investigate the problem of generalizing the router to unlearned selection space, we randomly selecte 7,000 queries from the previously mentioned 32,626 queries. These queries are sent to three state-of-the-art commercial VLMs: GPT-4o \cite{gpt4o}, Claude 3.5 Sonnet \cite{claude}, and Gemini 1.5 Flash \cite{gemini}, resulting in a total of 21,000 unique samples (7,000 queries × 3 VLMs). Fig.~\ref{fig:sample} shows an sample from M$^2$, which includes a text-image pair, the requested VLM, the output generated by the VLM, and additional metadata such as language, task, ground truth answer, and label. A label is marked as True if the VLM’s output exactly matches the ground truth answer. For the commercial VLMs that are accessed via paid APIs, we also record the cost and latency of each request, following the practice of HELM \cite{HELM} and VHELM \cite{VHELM}.

To evaluate the router’s performance in an out-of-distribution (OOD) scenario, we randomly select 10,000 queries from another benchmark (VHELM \cite{VHELM}). These queries are used to query both open-source and commercial VLMs, resulting in 7,084 unique queries that can be correctly answered by at least one free VLM, and
9,397 unique queries that can be correctly answered by at least one VLM.

\subsection{Data Analysis} 

The statistical information of M$^2$ is shown in Table.~\ref{tab:statistics}, which is divided into two scenarios: in-distribution (ID) and out-of-distribution (OOD). In the ID scenario, the samples are used to train and test the router's performance on in-distribution data. The VLM selection space size being either 4 or 7, totaling 35,126 unique image-text queries in the training set and 4,500 queries in the test set. In the OOD scenario, the samples are used to test the router's performance on out-of-distribution data. The VLM selection space size is either 4 or 7, totaling 16,481 unique image-text queries.

\begin{table}[!ht]
  \centering
  \caption{Dataset statistics}
    \begin{tabular}{cccrr}
    \toprule
    \multicolumn{3}{c}{VLM Pool} & \multicolumn{1}{l}{Training Set} & \multicolumn{1}{l}{Test Set} \\
    \midrule
    \multirow{2}[0]{*}{ID} & Free & \multicolumn{1}{l}{$|\mathcal{S}|$=4} & 29,626 & 3,000 \\
    & Free+Paid & \multicolumn{1}{l}{$|\mathcal{S}|$=7} & 5,500  & 1,500 \\
    \midrule
    \multirow{2}[0]{*}{OOD} & Free & \multicolumn{1}{l}{$|\mathcal{S}|$=4} & 0 & 7084 \\
    & Free+Paid & \multicolumn{1}{l}{$|\mathcal{S}|$=7} & 0 & 9397 \\
    \midrule
    \multicolumn{3}{c}{Total} & 35126 & 20981 \\
    \toprule
    \end{tabular}%
    \begin{tablenotes}
    \footnotesize
    \item `Free` denotes 4 open-sourced VLMs, while `Paid` denotes 3 commercial VLMs.
  \end{tablenotes}
  \label{tab:statistics}%
\end{table}%


\section{Router for Model Selection}
\label{methods}

\subsection{Problem Formalization}
The goal of a VLM router is to identify the cheapest and lowest-latency model that can correctly answer a given query. However, in our setting, the selection space $\mathcal{S}$ consists of four open-source VLMs that can be considered free to use. Moreover, inference latency varies significantly across hardware platforms, and we have no control over whether the hardware used in our experiments is consistent with that employed by the providers of the commercial VLMs in $\mathcal{S}$, making latency comparisons unreliable.

Therefore, we disregard cost and latency, and simplify the router’s objective to determining whether a given VLM can correctly answer a given query, which is an essential prerequisite for ensuring the actual performance of the router after query dispatch. It can be formalized as:

\begin{equation}
\max_{\Phi} \sum_{(Q_i, G_i)\in D} \sum_{M_j \in \mathcal{S}}  \log\left(
{p_{\Phi}
(y_{i,j}=\bar{y}_{i,j} | Q_i, G_i, M_j })
\right),
\label{equ:goal}
\end{equation}

where $Q_i$ denotes the question and $G_i$ denotes the image in the $i$-th query ($Q_i$, $G_i$), $D$ denotes the training set, and $M_j$ denotes profile of the $j$-th VLM from the selection space $\mathcal{S}$. $M_j$ is a sentence that contains the VLM's name, parameter count, architecture, as well as its language and vision models. $\Phi$ denotes the parameters of the router. For a given query $(Q_i, G_i)$, $y_{i,j}$ is the predicted label indicating whether $M_j$ is expected to answer the query correctly, and $\bar{y}_{i,j}$ is the corresponding ground-truth label. Specifically, $\bar{y}_{i,j} = 1$ if $M_j$ provides a correct answer to $(Q_i, G_i)$, and $\bar{y}_{i,j} = 0$ otherwise.

\subsection{Router Architecture}

\begin{figure}[!ht]
  \centering
  \includegraphics[width=\linewidth]{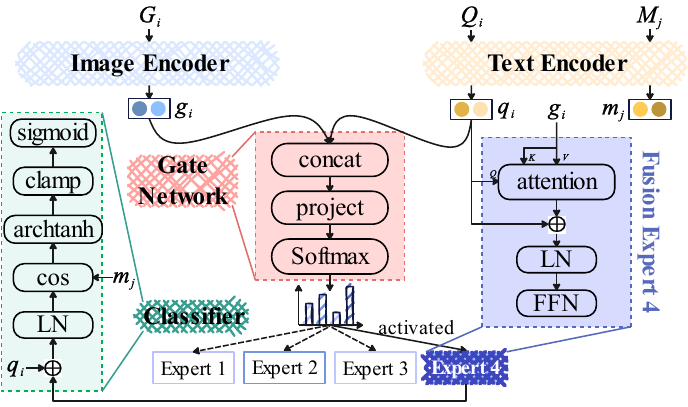}
  \caption{Architecture of ARMS, consists of five modules: a text encoder, an image encoder, a gate module, fusion experts, and a classifier.}
  \Description{Our method}
  \label{fig:ARMS}
\end{figure}

As shown in Fig.~\ref{fig:ARMS}, ARMS adopts a Mixture-of-Experts (MoE) architecture, comprising five modules: the text encoder, the image encoder, the gate network, the fusion experts, and the classifier. Based on Bert\cite{Bert}, the text encoder encodes the question $Q_i$ from the $i$-th multimodal query and the $j$-th VLM profile $M_j$ into vectors $q_i$ and $m_j$, respectively. Based on CLIP\cite{clip}, the image encoder encodes the image $G_i$ from the $i$-th multimodal query into the vector $g_i$.

\textbf{Gate network.} Since different VLMs have their own areas of specialization and user queries vary widely in domain, we employ the gate network that dynamically selects and activates expert modules based on the input query ($q_i$, $g_i$), enabling ARMS to effectively handle diverse tasks defined by both query domain and VLM specialization. It first concatenates the two input vectors, $q_i$ and $g_i$, then applies a projection followed by a softmax operation to obtain a probability distribution $p_i$, which can be formalized as:

\begin{equation}
p_i = \mathrm{softmax}\left(W [q_i; g_i] + b\right), 
\end{equation}
where $W$ denotes the weight matrix, $W \in \mathbb{R}^{N \times(d_q+d_g)}$, $b$ denotes the bias vector, $b \in \mathbb{R}^{N}$, and $N$ denotes the number of experts. $d_q$ and $d_g$ represent the dimensions of $q_i$ and $g_i$, respectively. $p_i$ is an $N$-dimensional vector, represents the probability of selecting each expert given the $i$-th input. Specifically, the probability of selecting the $i$-th expert is:

\begin{equation}
p_i^{(n)} = \frac{\exp\left(w_n^\top [q_i; g_i] + b_n\right)}{\sum_{k=1}^N \exp\left(w_k^\top [q_i; g_i] + b_k\right)}, \quad n = 1, 2, \dots, N
\end{equation}

\textbf{Fusion experts.} The query is multimodal, consisting of an image and a text question, the information embedded in the image is necessary to answer the text question. The fusion expert uses an attention mechanism to align the semantic representations of the image and text, facilitating the fusion of multimodal vectors. Each expert is a transformer block, which takes the question vector $q_i$ as the query and the image vector $g_i$ as both the key and value. It computes the attention values, concatenates them with $q_i$, and then passes the result through a Layer Normalization (LN) layer and a Feed-Forward Network (FFN) layer to obtain the fused vector $e_i$. It can be formalized as: 

\begin{equation}
e_i=\operatorname{FFN}\left(\operatorname{LN}\left(\left[q_i ; \operatorname{softmax}\left(\frac{q_i g_i^{\top}}{\sqrt{d}}\right) g_i\right]\right)\right), 
\end{equation}
where $\sqrt{d_q}$ is the scaling factor to prevents the dot product values from becoming too large.

\textbf{Classifier.} After multimodal vector fusion, the classifier determines whether the $j$-th VLM can correctly answer the $i$-th query based on the fused vector $e_i$ and the embedding vector of the VLM profile $m_j$. Specifically, it first concatenates $e_i$ with $q_i$, applies a LN layer, and then computes the cosine similarity with $m_j$, to obtain the score $s_{i,j}$, which can be formalized as:

\begin{equation}
s_{i, j}==\frac{\operatorname{LN}([q_i; e_i])^{\top} m_j}{\left\|\operatorname{LN}([q_i; e_i])\right\| \cdot\left\|m_j\right\|}
\end{equation}

The value of $s_{i,j}$ ranges from \([-1, 1]\), whereas the input to the cross-entropy loss function is expected to be logits in the range $(-\infty, +\infty)$. Therefore, we apply the inverse hyperbolic tangent $\mathrm{arctanh}$ to $s_{i,j}$ to obtain a temporary vector $t_{i,j}$. To prevent gradient explosion, we apply an interval projection on $t_{i,j}$, as shown in Eq.~\eqref{equ:clamp}, constraining each element of the input tensor within a predefined range \([\text{min}, \text{max}]\).

\begin{equation}
\text{clamp}(t_{i,j}) =
\begin{cases}
a, & \text{if } t_{i,j} < \text{min} \\
t_{i,j}, & \text{if } \text{min} \leq t_{i,j} \leq \text{max} \\
b, & \text{if } t_{i,j} > \text{max}
\end{cases}
\label{equ:clamp}
\end{equation}

Finally, we apply the sigmoid function to $\text{clamp}(t_{i,j})$ to map it to the $[0, 1]$ range, which serves as the confidence score $c_{i,j}$, representing ARMS' belief in the correct prediction. If $c_{i,j}$>0.5, we set $y_{i,j}$=1, otherwise, we set $y_{i,j}$=0.

\subsection{Training Strategy for Adaptation}
To explore how to adapt the router to a set of VLMs that were unseen during training, we propose two extension training strategies: Incremental Training and Independent Training. The core idea of incremental training is to update the parameters of the original router using samples from the newly added VLMs. Independent training, on the other hand, involves training two separate routers for the original VLM set and the newly added VLM set, respectively, and making joint decisions through a threshold‑based cascaded system. The following sections will detail these two strategies respectively.

\textbf{Incremental Training Strategy.} 
Given a router $\Phi$ trained on dataset $D_\mathcal{S}$ and a set of unseen VLM in $\mathcal{S}$ related samples $D_{\bar{\mathcal{S}}}$, the core idea of the incremental training strategy is to continue training $\Phi$ on $D_{\bar{\mathcal{S}}}$. It can be formalized as:

\begin{equation}
\max_{\Phi \sim \Delta \Phi } \sum_{(Q_i, G_i)\in D_{\bar{\mathcal{S}}}} \sum_{M_j \in \bar{\mathcal{S}}}  \log\left(
{p_{\sim \Delta \Phi}
(y_{i,j}=\bar{y}_{i,j} | Q_i, G_i, M_j })
\right),
\end{equation}
where $\sim$ denotes updating some parameters of $\Phi$, $\bar{\mathcal{S}}$ denotes the extended selection space, $|\bar{\mathcal{S}}|$>$|\mathcal{S}|$. $\Phi$ denotes the router obtained by initially training on $D_\mathcal{S}$ according to Eq.~\eqref{equ:goal}.

In our settings, $\mathcal{S}$=[Qwen2.5-VL-7B-Instruct, Idefics2-8B, Llama 3.2-11B-Vision-Instruct, SmolVLM2-2B-Instruct], $\bar{\mathcal{S}}$ - $\mathcal{S}$ = [GPT-4o, Claude 3.5 Sonnet, Gemini 1.5 Flash]. All samples in $D_\mathcal{S}$ are generated by one of the VLMs from the selection space $\mathcal{S}$, we use $\mathcal{S}$=4 to refer to this situation in the experiment (Sec.\ref{rq2}).

\textbf{Independent Training Strategy.} 
The independent training strategy aims to train two separate router, $\Phi$ and $\Theta$ on $D_\mathcal{S}$ and $D_{\bar{\mathcal{S}}-\mathcal{S}}$, respectively. When applied to different datasets, the training of each router can be uniformly formalized by Eq~\eqref{equ:goal}. All samples in $D_{\bar{\mathcal{S}}-\mathcal{S}}$ are generated by one of the VLMs from the new selection space $\mathcal{S}_{new}$ = $\bar{\mathcal{S}}$ - $\mathcal{S}$.

\subsection{Model Selection Process} 
Finally, we construct a threshold-based ($\theta$) cascaded system using the two trained routers ($\Phi$ and $\Theta$) to choose the VLM with high confidence score from the expanded selection space $\bar{\mathcal{S}}$. The selection process is illustrated in Algorithm.~\ref{alg:selection_process}. Given a query, the system first uses $\Phi$ to compute the confidence scores of all VLMs in $\mathcal{S}$, recording the highest confidence score $c_{max}$ and its corresponding VLM  $m_{\text{best}}$. If $c_{max}$ > $\theta$, the system invokes $m_{\text{best}}$ to answer the user's query. Otherwise, it uses $\Theta$ to compute the confidence scores of all VLMs in $\mathcal{S}_{new}$, and compares them with $c_{max}$, selecting the VLM with the higher confidence to answer the query.

\renewcommand{\algorithmicrequire}{\textbf{Input:}}
\renewcommand{\algorithmicensure}{\textbf{Output:}}
\begin{algorithm}[!t]
\caption{Selection Process for VLM Choice}
\label{alg:selection_process}
\begin{algorithmic}[1]
    \REQUIRE Query \( q \), threshold \( \theta \), original VLM set \( \mathcal{S} \), expanded VLM set \( \mathcal{S}_{new} \), routers \( \Phi \) and \( \Theta \)
    \ENSURE Selected VLM \( m_{\text{best}} \) for answering the query
    
    \STATE Use \( \Phi \) to compute the confidence scores for all VLMs in \( \mathcal{S} \)
    \STATE Record the highest confidence score \( c_{\text{max}} \) and its corresponding VLM \( m_{\text{best}} \)
    
    \IF{$c_{\text{max}} > \theta$}
        \STATE Invoke \( m_{\text{best}} \) to answer the user's query
    \ELSE
        \STATE Use \( \Theta \) to compute the confidence scores for all VLMs in \( \mathcal{S}_{new} \)
        \STATE Compare the confidence scores with \( c_{\text{max}} \) and select the VLM with the higher confidence
        \STATE Invoke the selected VLM to answer the user's query
    \ENDIF
\end{algorithmic}
\end{algorithm}

\begin{figure*}[!ht]
    \centering
    \begin{subfigure}{0.33\textwidth}
        \centering
        \includegraphics[width=\textwidth]{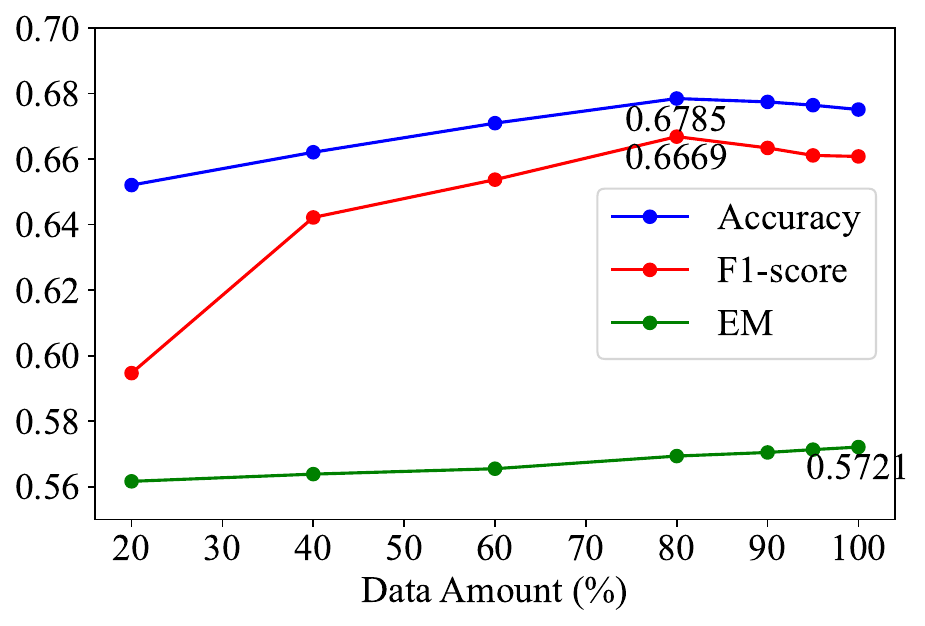}
        \caption{Impact of data amount on results}
        \Description{Overall}
        \label{fig:sub1}
    \end{subfigure}
    \hfill
    \begin{subfigure}{0.33\textwidth}
        \centering
        \includegraphics[width=\textwidth]{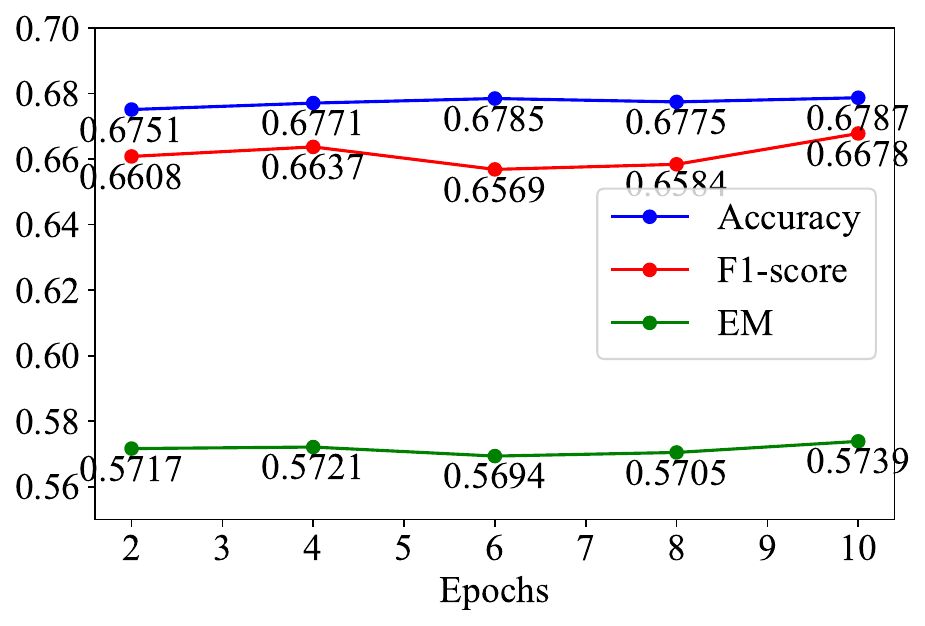}
        \caption{Impact of epochs on results, ID scenario}
        \Description{Ask}
        \label{fig:sub2}
    \end{subfigure}
    \hfill
    \begin{subfigure}{0.325\textwidth}
        \centering
        \includegraphics[width=\textwidth]{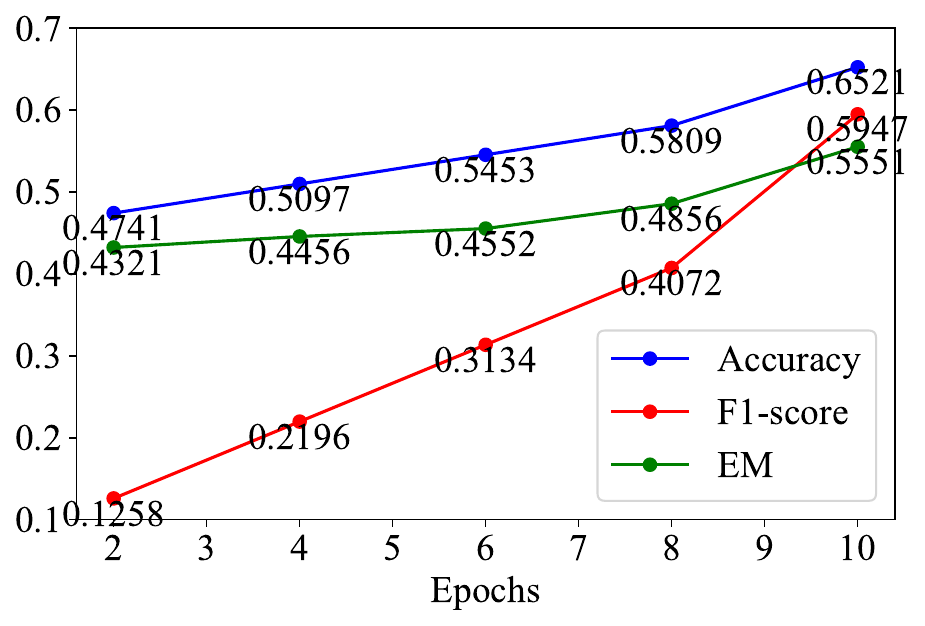}
        \caption{Impact of epochs on results, OOD scenario}
        \Description{Tell}
        \label{fig:sub3}
    \end{subfigure}
    \vspace{-3mm}
    \caption{Impact of data amount and training epochs on the performance of ARMS in in-distribution (ID) and
     out-of-distribution (OOD) scenarios. The best results are annotated below the corresponding points.}
    \label{MedDG-Derivations}
\end{figure*}

\section{Experiment}

\subsection{Experiment Setup}

\subsubsection{Research Questions} 

We conducted experiments to assess the effectiveness of our work by addressing the following research questions (RQ):

\textbf{\hyperref[rq1]{RQ1.}} 
How effective is the constructed dataset \textbf{M$^2$} in providing supervision signals for VLM selection? Specifically, can it improve both in-distribution performance and generalization to out-of-distribution scenarios as the training process scales?

\textbf{\hyperref[rq2]{RQ2.}} How does ARMS perform on the in-distribution and out-of-distribution test sets compared to individual VLMs? Specifically, what are the EM scores achieved by the regular ARMS and the extended version of ARMS on unseen VLMs?


\textbf{\hyperref[rq3]{RQ3.}} What is the impact of different modules on the performance of ARMS, and why is ARMS effective? Specifically, how can ARMS enhance the representations of image-text query and VLM capability?



\textbf{\hyperref[rq4]{RQ4.}} How do the proposed training strategies enable efficient and scalable adaptation to newly introduced VLMs under different scenarios?


\subsubsection{Evaluation Metrics}

We use Accuracy, Precision, Recall, and F1-Score to evaluate the \textbf{selection performance} of ARMS in the VLM selection problem. Followed by HELM \cite{HELM}, we use EM (Exact Match) to evaluate the \textbf{actual performance} of VLMs selected by ARMS on the test set, and EM=$\mathcal{N}/\mathcal{M}$, where $\mathcal{N}$ represents the number of queries for which the selected VLM provides the exact correct answer, and $\mathcal{M}$ is the total number of queries in the test set.

\begin{figure*}[!ht]
    \centering
    \begin{subfigure}{0.40\textwidth}
        \centering
        \includegraphics[width=\textwidth]{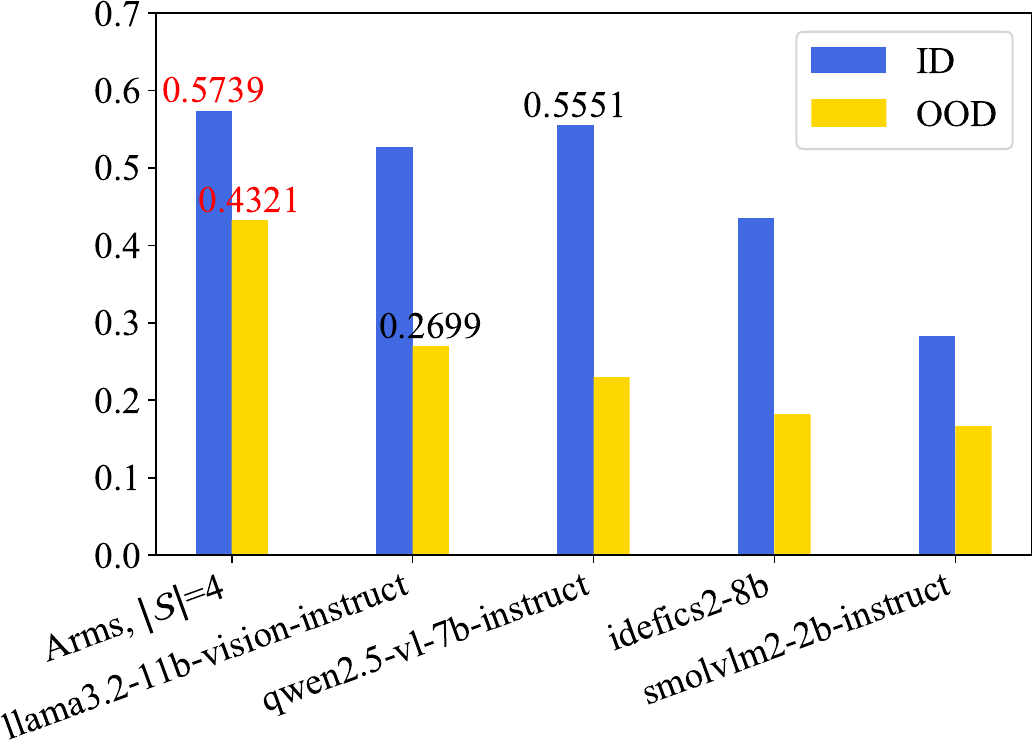}
        \caption{EM Results for $|\mathcal{S}|$=4, epochs=2}
        \Description{Overall}
        \label{fig:suba}
    \end{subfigure}
    \hfill
    \begin{subfigure}{0.58\textwidth}
        \centering
        \includegraphics[width=\textwidth]{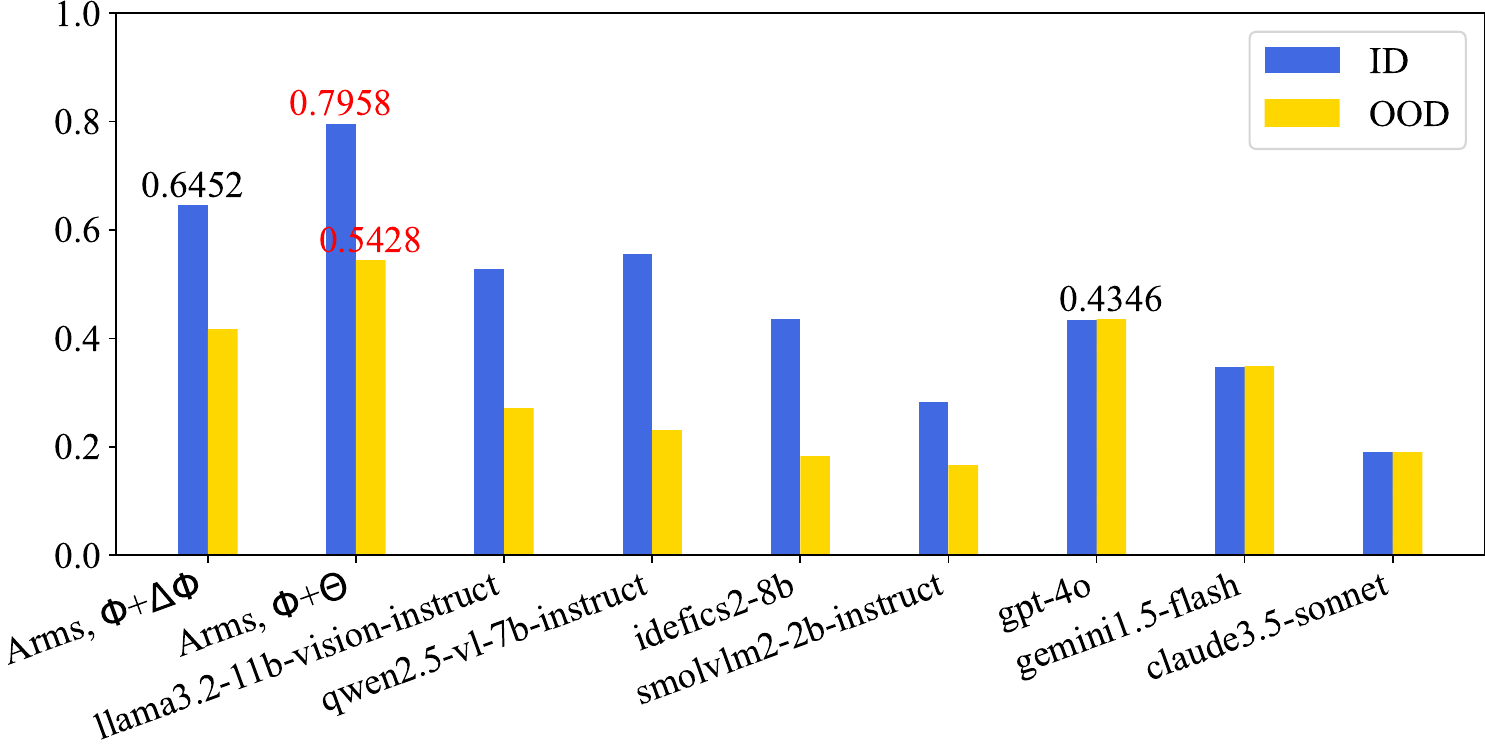}
        \caption{EM Results for $|\mathcal{S}|$=7, epochs=2}
        \Description{Ask}
        \label{fig:subb}
    \end{subfigure}
    \caption{Main Results of our router ARMS with different training strategies and selection spaces ($\mathcal{S}$). $\Phi$+$\Delta\Phi$ denotes the incremental training strategy and $\Phi$+$\Theta$ denotes the independent training strategy. The best and the second-best results are annotated above the corresponding bars, the best results are in \textcolor{red}{red}. $|\mathcal{S}|=4$ denotes that $\mathcal{S}$ contains only four open-source VLMs, while $|\mathcal{S}|=7$ means that $\mathcal{S}$ includes four open-source models and three commercial VLMs}
    \label{fig:Main}
\end{figure*}

\subsubsection{Baselines and Oracle Results}

As there is currently no prior work specifically targeting VLM selection, and existing LLM selection approaches are not directly applicable to the multimodal setting considered in this paper, we treat each of the seven individual
VLMs in the selection space of ARMS as baselines for comparison. These include GPT-4o \cite{gpt4o}, Claude 3.5 Sonnet \cite{claude}, Gemini 1.5 Flash, Qwen2.5-VL-7B-Instruct \cite{qwen2.5}, Idefics2-8B \cite{idefics2}, Llama 3.2-11B-Vision-Instruct \cite{llama3.2}, and SmolVLM2-2B-Instruct \cite{smolvlm2}. 
Followed by FORC \cite{FORC}, we use the oracle result as the performance upper bound of the router by always choosing the VLM that solves the query. In our dataset, the oracle EM result is 1.0, as only queries for which at least one VLM provides a correct answer are retained, ensuring a balanced distribution of positive and negative samples.

\subsubsection{Implementation Details}
The hyper-parameters are configured with a batch size of 64, a learning rate of 8e-5, a weight decay of 1e-2 and we use the AdamW \cite{AdamW} optimizer. ARMS uses CLIP-ViT-B/32 \cite{clip} as image encoder and uses DistilBERT \cite{DistilBERT} as the text encoder. We set the expert number $N$ is 4, set the dimension of vision vector $g_i$ is 512 and the dimensions of textual vector
$q_i$ and $m_j$ are both 768. All experiments are conducted using 4 NVIDIA RTX 4090 GPUs and a 16-core Xeon(R) Gold 6430 CPU.

\subsection{Impact of Data Amount (RQ1)}
\label{rq1}

\subsubsection{Answer to RQ1} 

The results show that ARMS benefits consistently from the supervision provided by \textbf{M$^2$}. As the amount of training data increases, the EM score improves steadily, indicating that \textbf{M$^2$} provides increasingly reliable and informative signals for model selection. Although Accuracy and F1 peak at a certain point and then slightly decline, the continuous improvement in EM suggests that larger-scale data helps refine more precise selection decisions. Under the in-distribution (ID) setting, increasing the number of training epochs brings only marginal performance gains, even when using the full training data. In contrast, under the out-of-distribution (OOD) setting, additional training epochs significantly improve performance. This demonstrates that \textbf{M$^2$} not only supports effective learning within seen distributions but also enables ARMS to acquire more generalized and transferable selection patterns, leading to substantial gains in OOD scenarios.


\subsubsection{Analysis}
As shown in Fig.~\ref{fig:sub1}, the accuracy and F1 score of ARMS increase with the training data amount and peak at 80\% of the training data, while the EM score continues to improve consistently as more data is introduced. This observation suggests that M$^2$ provides increasingly stable and informative supervision signals for model selection, where additional data helps refine fine-grained decision boundaries that are better reflected by the EM metric. As shown in Fig.~\ref{fig:sub2}, increasing the number of training epochs has only a marginal impact on the ID test set, with the EM score rising slightly from 0.5717 to 0.5739. This indicates that ARMS can already effectively capture selection patterns within the seen distribution with limited optimization. In contrast, Fig.~\ref{fig:sub3} shows that additional training epochs significantly improve performance on the OOD test set, boosting the EM score from 0.4321 to 0.5551 (a 28.44\% increase).

This substantial gain demonstrates that M$^2$ not only supports fitting to in-distribution data but also enables ARMS to learn more generalized and transferable multimodal alignment patterns. As training proceeds, the model gradually acquires more robust selection criteria that extend beyond seen data, highlighting the critical role of M$^2$ in improving generalization for VLM selection.

\begin{figure*}[!t]
    \centering
    \begin{subfigure}{\columnwidth}
        \centering
        \includegraphics[width=0.95\textwidth]{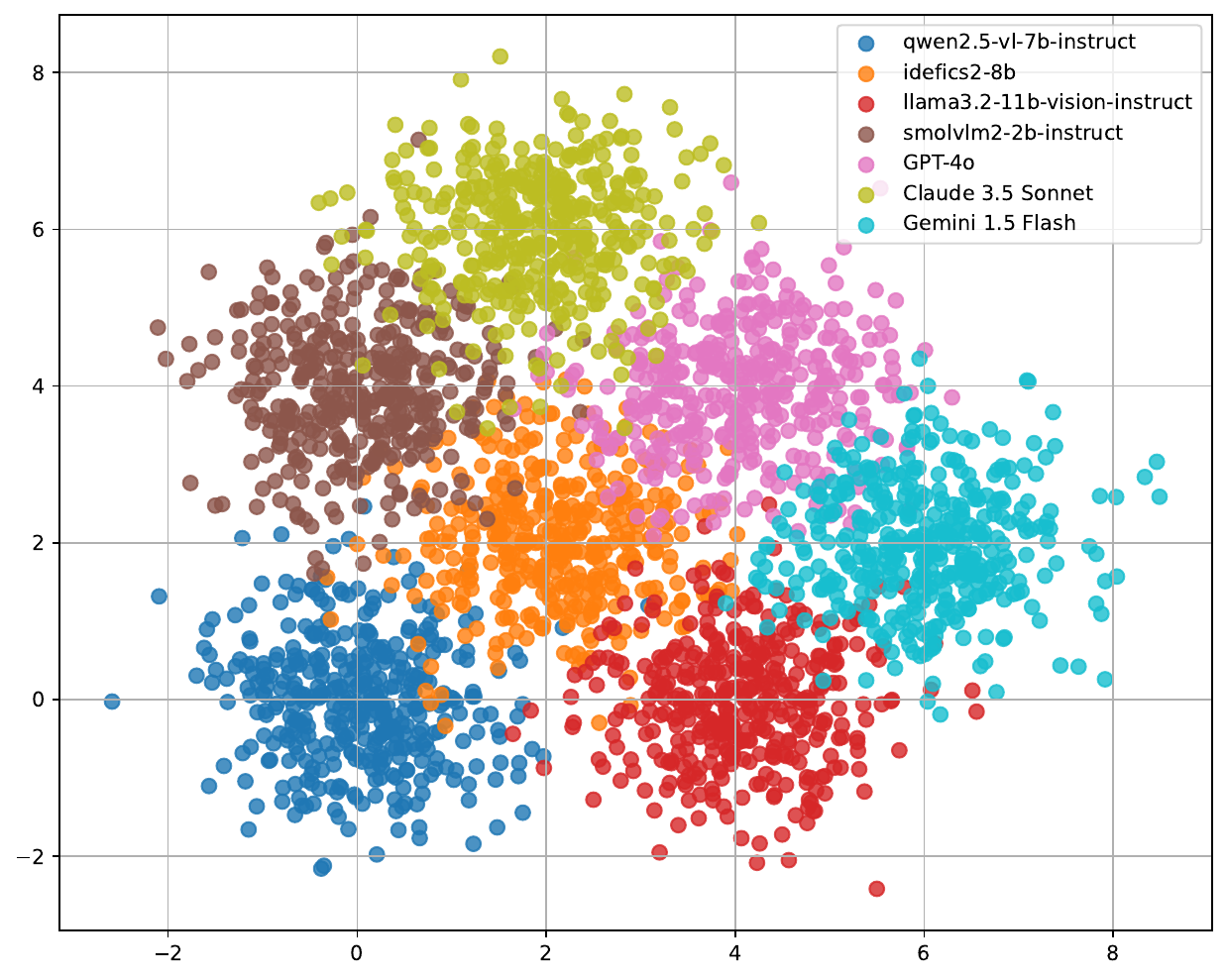}
        \caption{Queries in different VLMs with capabilities are color-coded}
        \Description{Capabilities}
        \label{fig:VLM_clusters}
    \end{subfigure}
    \hfill
    \begin{subfigure}{\columnwidth}
        \centering
        \includegraphics[width=0.95\textwidth]{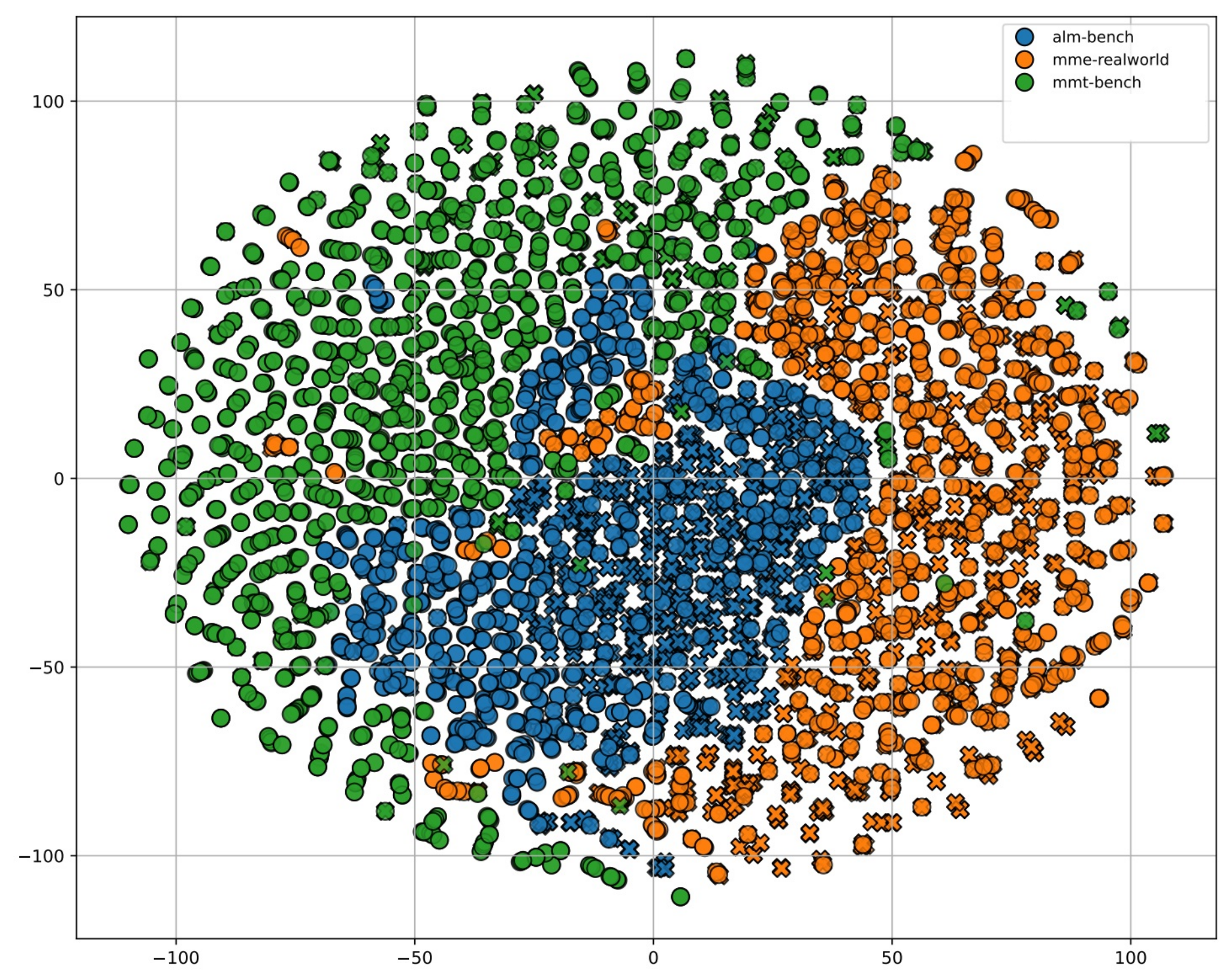}
        \caption{Queries in different benchmarks are color-coded.}
        \Description{Difficulty}
        \label{fig:visual}
    \end{subfigure}
    \vspace{-2mm}
    \caption{Embeddings visualization of the fused image and text vectors from ARMS with t-SNE \cite{SNE}.}
    \label{fig:rq4}
\end{figure*}

\subsection{Main Results (RQ2)}
\label{rq2}

\subsubsection{Answer to RQ2} Under both ID and OOD settings, ARMS consistently outperforms all four open-source individual VLMs in EM score. Furthermore, when three SOTA commercial VLMs are added to the selection space $\mathcal{S}$ using our two training strategies, ARMS surpass any individual VLM in the extended set $\mathcal{S}$. This highlights the scalability of ARMS and its ability to generalize across diverse VLM combinations.

\subsubsection{Analysis} 

As shown in Fig.~\ref{fig:suba}, when the selection space $\mathcal{S}$ contains only four open-source models, ARMS achieves the best performance among them under both ID and OOD settings, with EM scores of 0.5739 and 0.4321 respectively, despite being trained for only two epochs. This highlights the value of using a router to select an appropriate VLM to answer user queries.

As shown in Fig.~\ref{fig:subb}, after incorporating a new commercial model into $\mathcal{S}$ using the independent training strategy, ARMS further improves its performance, achieving the best results under both ID and OOD settings—0.7958 and 0.5428, respectively. However, there remains a significant gap from the upper bound of 1.0 (oracle result), indicating that our M$^2$ dataset poses a highly challenging task.

Moreover, the incremental training strategy performs worse than the independent strategy under the ID setting and worse than GPT-4o under the OOD setting. This is because, under our experimental setup, the number of training samples for the newly added VLMs is much smaller than that for the original VLMs in $\mathcal{S}$, making it difficult for the router to learn effectively in a low-resource regime. This suggests that when expanding the router's selection space, a more effective approach under limited data conditions is to train a separate router on the new VLMs and combine it with the original router to form a threshold-based cascaded system.

\subsection{Ablation Study (RQ3)}
\label{rq3}
\subsubsection{Answer to RQ3}
Each module in ARMS contributes positively to the final performance, with Bert yielding the most substantial gain (0.5721 vs. 0.3480). In addition, the attention-based fusion mechanism surpasses simple concatenation ($\oplus$), demonstrating the effectiveness of the attention mechanism introduced in ARMS for modeling different queries. The performance with VLM profiles surpasses that without profiles, demonstrating that the information in VLM profiles helps enhance the input signals. 

\begin{table}[!t]
  \centering
  \caption{Ablation study of each module in ARMS. {\color{cyan}\faSnowflake} denotes freezing the module parameters, while {\color{orange}\faFire} denotes updating the module. $\oplus$ denotes directly concatenating the embedding vectors of the two modules. w/o denotes that the VLM profiles are not used, and only the VLM name is used as $M_j$. All variants of ARMS are trained using the training set from the free VLM pool shown in Table~\ref{tab:statistics}.
  }
    \begin{tabular}{lccccc}
    \toprule
    Method & Acc. & Precision &Recall & F1 & EM \\
    \midrule
    CLIP{\color{cyan}\faSnowflake}~Bert{\color{orange}\faFire} & 0.6542 & 0.6270 & 0.6644 & 0.6452 & 0.5351 \\
    CLIP{\color{orange}\faFire}~Bert{\color{cyan}\faSnowflake} & 0.5875 & 0.3558 & 0.6630 & 0.4631 & 0.3480\\
    CLIP $\oplus$ Bert & 0.6577 & 0.6091 & 0.5979 & 0.6036 & 0.5272 \\
    w/o profiles 
   & 0.6254 & 0.5994 & 0.6351 & 0.6168 & 0.5115
     \\
    ARMS  & \textbf{0.6751} & 
    \textbf{0.6403} & \textbf{0.6882} & \textbf{0.6634} & \textbf{0.5721} \\
    \bottomrule
    \end{tabular}%
  \label{tab:Ablation}%
\end{table}%

\begin{figure*}[!ht]
    \centering
    \begin{subfigure}{\columnwidth}
        \centering
        \includegraphics[width=\textwidth]{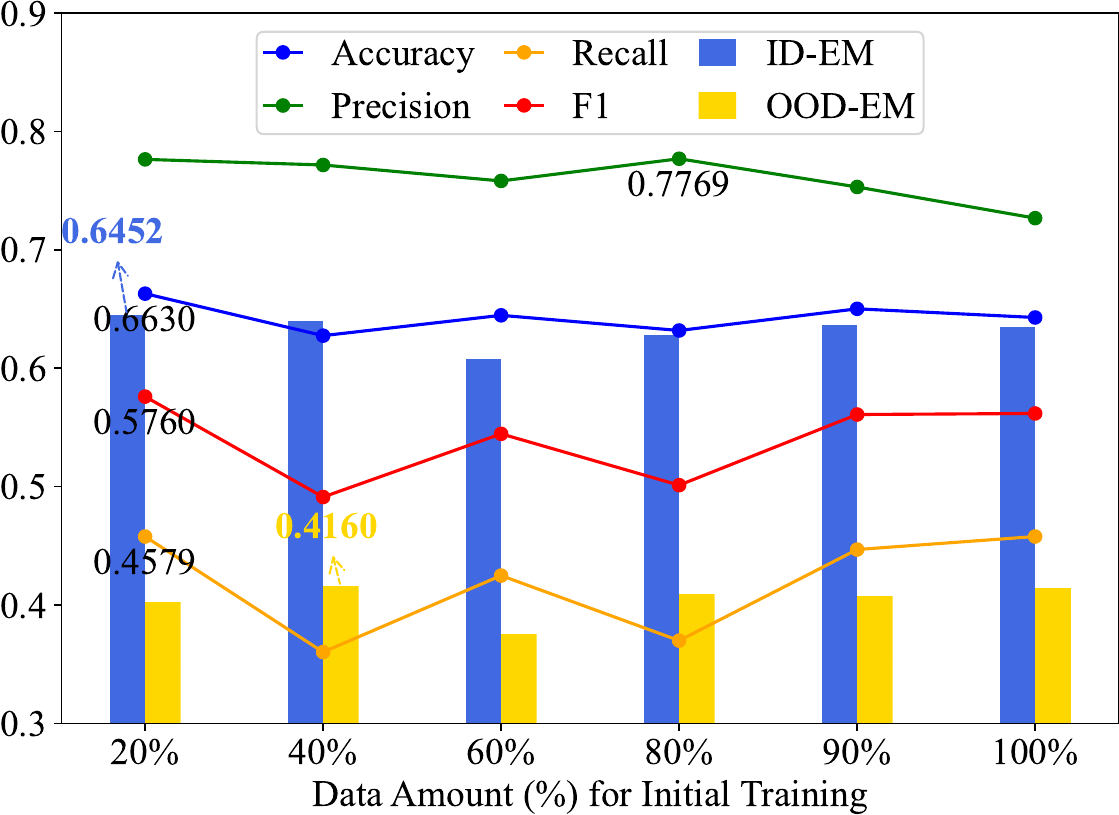}
        \caption{Incremental training strategy}
        \Description{Incremental}
        \label{fig:Incremental}
    \end{subfigure}
    \hfill
    \begin{subfigure}{\columnwidth}
        \centering
        \includegraphics[width=\textwidth]{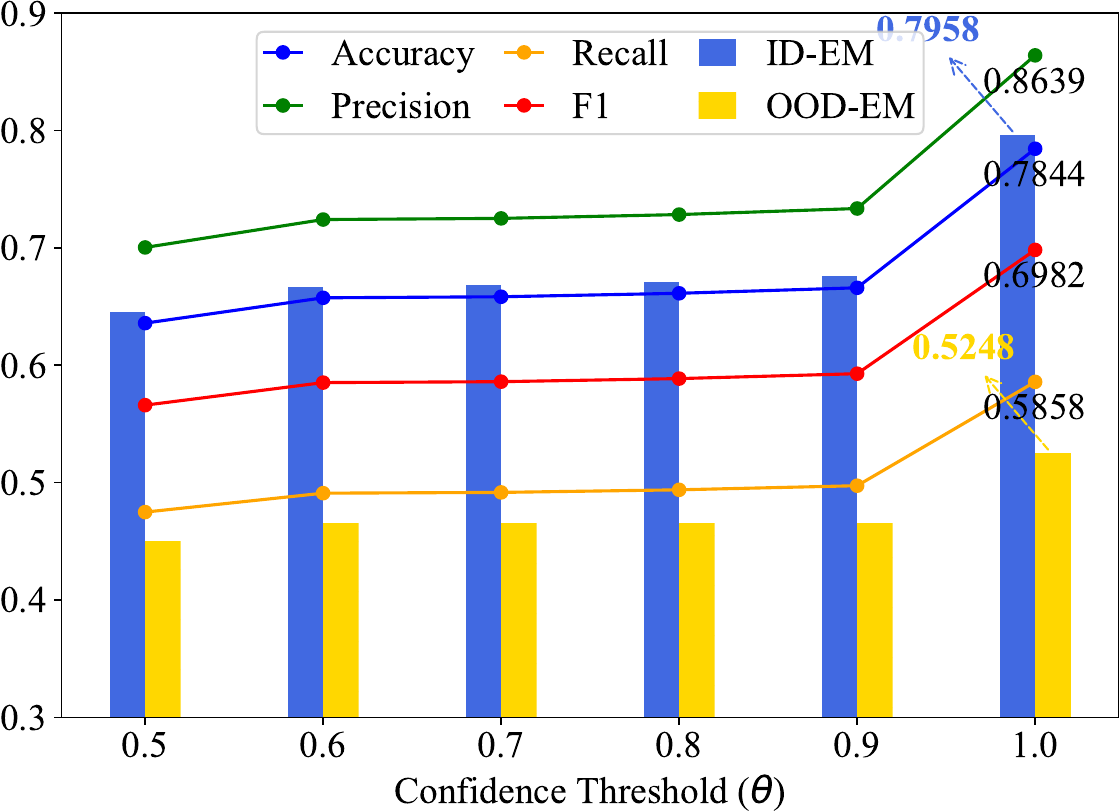}
        \caption{Independent training strategy}
        \Description{Independent}
        \label{fig:Independent}
    \end{subfigure}
    \caption{Comparison between two training strategies. The best results on Accuracy, Recall, Precision and F1-score are annotated below the corresponding points. The best results on EM are annotated above the corresponding bars.}
    \label{fig:twoS}
\end{figure*}

Using t-SNE to visualize the fused image and text vectors from ARMS in a low-dimensional space reveals clear cluster structures. These clusters correspond to representations of different queries or VLM capabilities. Vectors within each cluster are closely packed, while distinct clusters are well separated, indicating that the ARMS learns to capture the nonlinear behaviors of VLMs with different capabilities on queries, enabling more accurate routing.

\subsubsection{Analysis}
As shown in Table~\ref{tab:Ablation}, freezing BERT while updating only CLIP causes a significant decline in precision, ultimately lowering the EM score. This highlights the importance of textual encoders in ARMS. When the text and image vectors in the query are simply concatenated by $\oplus$, ARMS shows a substantial decline in recall (0.6882 $\rightarrow$ 0.5979). This indicates that improper multimodal fusion may lead to over-conservative decision boundaries, where the model prefers to predict positives only under high confidence, thus sacrificing recall to minimize false positives. However, our proposed attention-based fusion method can make more informed and balanced predictions, improving recall (0.6882) without significantly compromising precision (0.6403). ARMS uses VLM profiles outperforms the variant without profiles (w/o profiles) across all metrics. Specifically, EM improves from 0.5115 to 0.5721, and F1 increases from 0.6168 to 0.6634. This demonstrates that VLM profiles provide additional, effective signals that help ARMS more accurately determine whether a VLM can correctly answer a given query.

ARMS achieves strong performance by leveraging a Mixture-of-Experts (MoE) framework to dynamically select appropriate multimodal fusion experts based on each input query. This design allows ARMS to effectively learn the interaction between different queries and VLM capabilities. As shown in Fig~\ref{fig:VLM_clusters}, when coloring the query embeddings by benchmark, clear cluster structures emerge, indicating that ARMS effectively distinguishes representation of different queries across multiple benchmarks. Similarly, in Fig~\ref{fig:visual}, when coloring the fused representations by VLM, distinct clusters appear corresponding to VLMs with different capability ranges. Each cluster reflects a specific range of VLM capability, demonstrating that the MoE experts successfully specialize in modeling nonlinear differences in model performance. Together, these visualizations provide strong evidence that ARMS’ MoE design enhances the representations of both query and VLM capability, supporting its superior routing performance.

\subsection{Training Strategies Comparison (RQ4)}
\label{rq4}

\subsubsection{Answer to RQ4}

The results show that the two training strategies exhibit complementary strengths under different scenarios. Incremental training is more suitable when the selection space evolves gradually and prior knowledge can be reused, while independent training is more robust to distribution shifts and better suited for scenarios requiring high-confidence decisions. These findings demonstrate that ARMS enables flexible and e ff adaptation without requiring full retraining, supporting scalable extension to new VLMs.

\subsubsection{Analysis} 

The results reveal that the two training strategies exhibit distinct and complementary behaviors under different settings. As shown in Fig.~\ref{fig:Incremental}, the performance of incremental training depends on the balance between the initial training data and the newly introduced data. When the initial training data is relatively limited (e.g., 20\%), incremental training achieves its best performance under the ID setting, indicating its effectiveness in scenarios where the model space evolves gradually and prior knowledge can be effectively reused. However, as the amount of initial training data increases, the relative proportion of newly introduced data decreases, leading to a degradation in performance. This suggests that incremental training is sensitive to data imbalance across training stages and is better suited for settings where new models are introduced progressively with sufficient adaptation data.

In contrast, the independent training strategy demonstrates stronger robustness, particularly under OOD settings. As shown in Fig.~\ref{fig:sub1}, independent training consistently achieves higher performance than incremental training when the selection space is expanded, with significant improvements observed under both ID and OOD settings. This indicates that separating the training process for different model subsets effectively mitigates interference between heterogeneous VLMs and improves adaptation to distribution shifts.

Furthermore, Fig.~\ref{fig:Independent} shows that the performance of independent training remains stable across a wide range of confidence thresholds ($\theta$ from 0.5 to 0.9), suggesting that the router maintains reliable discriminative capability under varying decision criteria. Notably, when $\theta$=1.0, all performance metrics improve substantially, indicating that enforcing stricter confidence requirements can further enhance selection reliability. 

Overall, these results demonstrate that incremental training and independent training serve different roles in enabling scalable adaptation: incremental training supports efficient extension when new VLMs are gradually introduced, while independent training provides a more robust solution for scenarios involving significant distribution shifts or heterogeneous model pools. Together, they offer a flexible and scalable framework for extending the model selection space without requiring full retraining.

\section{Conclusions}
In this work, we introduce \textbf{M$^2$}, a large-scale multimodal dataset for VLM selection, and propose \textbf{ARMS}, a router for VLMs. This paper addresses three key challenges in developing an effective VLM router system, namely, predicting whether a given VLM can correctly answer a given query. ARMS outperforms individual VLMs in both in-distribution (ID) and out-of-distribution (OOD) settings. The experimental results demonstrate that VLM profiles enhance input signals, the MoE architecture effectively learns the representaions of different queries and VLM capabilitis. Besides, the extension training strategy is proven to be useful when adapting router to new VLMs. We believe that our work provides an effective solution for improving the performance and deployment efficiency of VLM-based applications in real-world scenarios.

\printbibliography

\appendix

\end{document}